\documentclass[11pt]{scrartcl} 
\ifx\pdfminorversion\undefined
\else
\fi

\usepackage{fullpage,epsfig}
\usepackage{comment}
\usepackage{xcolor}
\usepackage{longtable}
\usepackage{multirow,array}
\usepackage{authblk}
\usepackage{parskip}
\PassOptionsToPackage{hyphens}{url}
\usepackage{hyperref}
\usepackage{color,xspace}
\usepackage{cancel}

\usepackage{enumitem}

\usepackage{amsmath,amsfonts}
\usepackage{algorithmic}
\usepackage{algorithm}
\usepackage{array}
\usepackage{textcomp}
\usepackage{stfloats}
\usepackage{url}
\usepackage{verbatim}
\usepackage{graphicx}
\usepackage{cite}

\usepackage{hyperref}
\usepackage[nameinlink]{cleveref}
\Crefname{figure}{{Fig.}}{{Figs.}}
\Crefname{table}{{Table.}}{{Tables.}}
\Crefname{equation}{{}}{{}}
\Crefname{Algorithm}{}{}
\usepackage[ruled,vlined,linesnumbered,algo2e]{algorithm2e}
\usepackage{booktabs}
\usepackage{xcolor}
\usepackage{lineno}
\usepackage{subfigure}

\newcommand{\remove}[1]{}

\title{Long-distance Geomagnetic Navigation in GNSS-denied Environments with Deep Reinforcement Learning}

\author{Wenqi Bai\footnote{W. Bai, X. Zhang, and S. Yang are with Xi'an University of Technology, China (bayouenqy@outlook.com, xhzhang@xaut.edu.cn, yang.son.nan@gmail.com).} \; Xiaohui Zhang$^\ast$ \; Shiliang Zhang\footnote{S. Zhang is with University of Oslo, Norway (shilianz@uio.no).} \; Songnan Yang$^\ast$ \\ Yushuai Li\footnote{Y. Li is with Aalborg University, Denmark (yushuaili@ieee.org).} \; Tingwen Huang\footnote{T. Huang is with Shenzhen University of Advanced Technology, China (huangtingwen2013@gmail.com).\\This work was partially supported by the National Major Scientific Instrument Development Project of China under Grant 62127809; and the Basic Research in Natural Science and Enterprise Joint Foundation of Shaanxi Province under Grant 2021JLM-58.}}

\begin{document}
\maketitle

\begin{abstract}
Geomagnetic navigation has drawn increasing attention with its capacity in navigating through complex environments and its independence from external navigation services like global navigation satellite systems (GNSS). Existing studies on geomagnetic navigation, \textit{i.e.}, matching navigation and bionic navigation, rely on pre-stored map or extensive searches, leading to limited applicability or reduced navigation efficiency in unexplored areas. To address the issues with geomagnetic navigation in areas where GNSS is unavailable, this paper develops a deep reinforcement learning (DRL)-based mechanism, especially for long-distance geomagnetic navigation. The designed mechanism trains an agent to learn and gain the magnetoreception capacity for geomagnetic navigation, rather than using any pre-stored map or extensive and expensive searching approaches. Particularly, we integrate the geomagnetic gradient-based parallel approach into geomagnetic navigation. This integration mitigates the over-exploration of the learning agent by adjusting the geomagnetic gradient, such that the obtained gradient is aligned towards the destination. We explore the effectiveness of the proposed approach via detailed numerical simulations, where we implement twin delayed deep deterministic policy gradient (TD3) in realizing the proposed approach. The results demonstrate that our approach outperforms existing metaheuristic and bionic navigation methods in long-distance missions under diverse navigation conditions.
\end{abstract}

\section{Introduction}
Global navigation satellite systems (GNSS) estimate the position of an object by measuring the time delay of signals, known as pseudoranges transmitted from multiple satellites. GNSS is widely adopted for long-range navigation with its provision of precise location information without cumulative error~\cite{hu2023performance,zhang2021prediction,qi2023geographic}, as it recalculates position continuously based on satellite signals. The global coverage of GNSS allows it to work reliably across diverse terrains without the need for local navigation infrastructure. However, in environments \textit{e.g.}, tunnels or dense forests, GNSS signals are highly susceptible to environmental disruptions that can lead to degraded signal strength or complete signal loss~\cite{wang2020novel,zhang2021adaptive}. Accurate GNSS positioning typically requires signals from at least four satellites. When fewer than four satellites are available, positioning becomes infeasible, resulting in GNSS outages \cite{chen2024error}. In underwater environments, GNSS signals are completely unavailable~\cite{ioannou2024underwater}, rendering it impossible to conduct long distance navigation using GNSS services.

Geomagnetic navigation has drawn increasing attention with its using the ubiquitous geomagnetic signals for navigation \cite{zhang2020geomagnetic,pang2023ins,liu2021multisurrogate}. Geomagnetic approaches are free from dependence on GNSS service or inertia navigation equipment and are with confined drift and cumulative errors \cite{pang2023ins,chen2023geomagnetic}, making it promising especially for long-distance navigation missions in unexplored areas or underwater missions where GNSS service is inaccessible. One of the most extensively investigated geomagnetic approaches carries out navigation via geomagnetic matching~\cite{wang2019triangle, xu2022innovative, wang2022geomagnetic}. Geomagnetic matching compares and matches real-time geomagnetic measurements with a geomagnetic map to determine the location based on matching result. However, geomagnetic matching requires a pre-established geomagnetic map with optimal integrity and high precision, which is impossible for navigation across unexplored areas.

Numbers of studies have shown that animals like pigeons and turtles can use the geomagnetic field for long-distance migration and homing\cite{zhou2022bionic, wiltschko2007magnetoreception, lohmann1996orientation, putman2015inherited, boles2003true}, without any pre-stored maps in their brain. Furthermore, researchers have found that there exists an almost one-to-one correlation between the geomagnetic field vector and a specific location in near-earth space \cite{langel1987main}. Inspired by the magnetic orientation behavior observed in animals, bionic geomagnetic navigation has gained momentum to derive the navigation based on real-time measured geomagnetic signals~\cite{DBLP:journals/corr/abs-2403-08808}. The problem of bionic geomagnetic navigation is framed as the autonomous exploration of a navigation path in response to the geomagnetic environment during the navigation \cite{zhang2020geomagnetic}.

Several bionic geomagnetic navigation approaches have been studied. Liu \textit{et al.}~\cite{liu2013bio} introduced a bio-inspired geomagnetic navigation method for unmanned vehicles. They characterized the navigation process as the convergence of geomagnetic measurements towards the destination, and proposed a stress evolution search method to achieve such convergence. Their experimental results demonstrate efficient navigation without any prior geomagnetic map. However, they use a random searching strategy for the navigation that leads to significant fluctuations in heading angles and zigzag path patterns. A segmented searching approach is proposed to optimize and expedite trajectory convergence \cite{guo2018bio}. This approach employs the probability evolution strategy (PES) for the initial navigation search and can compensate the random searching. Taylor \textit{et al.}~\cite{taylor2021long} explored the potential of magnetic inclination in guiding long-distance navigation. They employed an agent-based modeling approach and succeeded in trans-equatorial navigation missions using magnetic inclination measurements. Zhou \textit{et al.}~\cite{zhou2022bionic} proposed a differential evolution (DE) algorithm to improve the searching efficiency in geomagnetic navigation, where they use enhanced mutation in the evolution strategy to attain superior individuals for the searching. Qi \textit{et al.}~\cite{qi2023geographic} utilized geomagnetic gradients of real-time measurements to predict the heading direction towards the destination. One of the requirements of this approach is that the gradients of geomagnetic intensity and inclination are not parallel in the navigation areas to guarantee an efficient gradient calculation, while such conditions might not hold in navigation practice. Overall, bionic geomagnetic navigation approaches iteratively generate populations to explore a solution space, from where an optimal solution is expected to lead to successful navigation. Nevertheless, the exploration process inevitably involves the execution of suboptimal solutions, which renders high cost or even failures especially in long-range navigation, making it a typical expensive optimization problem~\cite{liu2021multisurrogate}.

Deep reinforcement learning (DRL) has been adopted to solve optimization problems in various applications, including visual navigation~\cite{lobos2018visual,jiang2023neuro} and radar navigation~\cite{zhang2022ipaprec,miranda2023generalization} where DRL trains agents to learn the optimal navigation strategy through interactions with the environment. Instead of random or heuristic searching, DRL emphasizes the role of environment in shaping the learning of agent towards the optimal solution. Through a learning process based on trial and error, DRL agents continuously refine and converge to the optimized solution without the need to repeatedly assess suboptimal choices~\cite{sutton2018reinforcement,guo2023optimal,zhang2023state}. The DRL features in optimization make DRL well-suited for long-distance geomagnetic navigation, where real-time decision-making is essential in unexplored navigation areas with unknown geomagnetic conditions. The key challenge in applying DRL to geomagnetic navigation lies in accurately modeling the complex and non-linear decision-making processes in the dynamic navigation environment, while sustaining the efficiency and adaptability of the DRL agent in navigation missions.

In this work, we propose a DRL-based approach for long-distance geomagnetic navigation in GNSS-denied and unexplored environments. Particularly, we adopt the DRL framework of twin delayed deep deterministic policy gradient (TD3), and design a geomagnetic gradient-guided TD3 (Gradient-Guided TD3) algorithm to learn the optimal navigation solution for long-distance missions. We leverage the continuous action space of TD3 to capture the subtle variations in the magnetic field during navigation, to maintain stable and efficient navigation progress. To the best of our knowledge, we are the first to develop a DRL-based approach for long-distance geomagnetic navigation. We summarize the contributions of our work as follows.

\begin{enumerate}
    \item We develop a DRL-based geomagnetic navigation strategy where we propose a Gradient-Guided TD3 algorithm, leveraging ubiquitous geomagnetic information of the earth for navigation decisions in long-range missions where GNSS service can be unavailable and the geomagnetic features unknown during the navigation.
    
    \item To guarantee the exploration efficiency of the DRL agent in navigation, we design a dense reward based on geomagnetic gradient information. This encourages the agent to choose smooth and efficient navigation actions and also improves generalizability of the trained DRL agent.

    \item We evaluate the proposed Gradient-Guided TD3 method in simulations under long-distance navigation scenarios with diverse conditions. We conduct comprehensive assessments, and compare our approach with existing heuristic geomagnetic navigation algorithms to demonstrate the performance of the developed approach.
\end{enumerate}

The remaining of this article is organized as follows. \Cref{Section:2} describes the fundamentals for geomagnetic navigation. \Cref{Section:3} details the proposed Gradient-Guided TD3 method. \Cref{Section:4} carries out simulations to compare the performance of Gradient-Guided TD3 with heuristic algorithms, and analyzes the results in details and demonstrates the effectiveness of our approach. We conclude this work in \Cref{Section:5}.

\section{Fundamentals}
\label{Section:2}

\subsection{Mathematical Description of the Geomagnetic Field}
\label{Section:2.1}

The geomagnetic field is a fundamental physical field of the earth, theoretically corresponding to the geographical position of any point in near-earth space \cite{tyren1982magnetic}. The strength of the geomagnetic field is typically measured in nanoteslas (nT). The average intensity on the ground is \(5\times 10 ^4\) nT, reaching only about \(7\times 10 ^4\) nT at the two poles, indicating that the geomagnetic field is considered a weak magnetic field.

\begin{figure}[htb]
    \centering
    \includegraphics[width=0.5\linewidth]{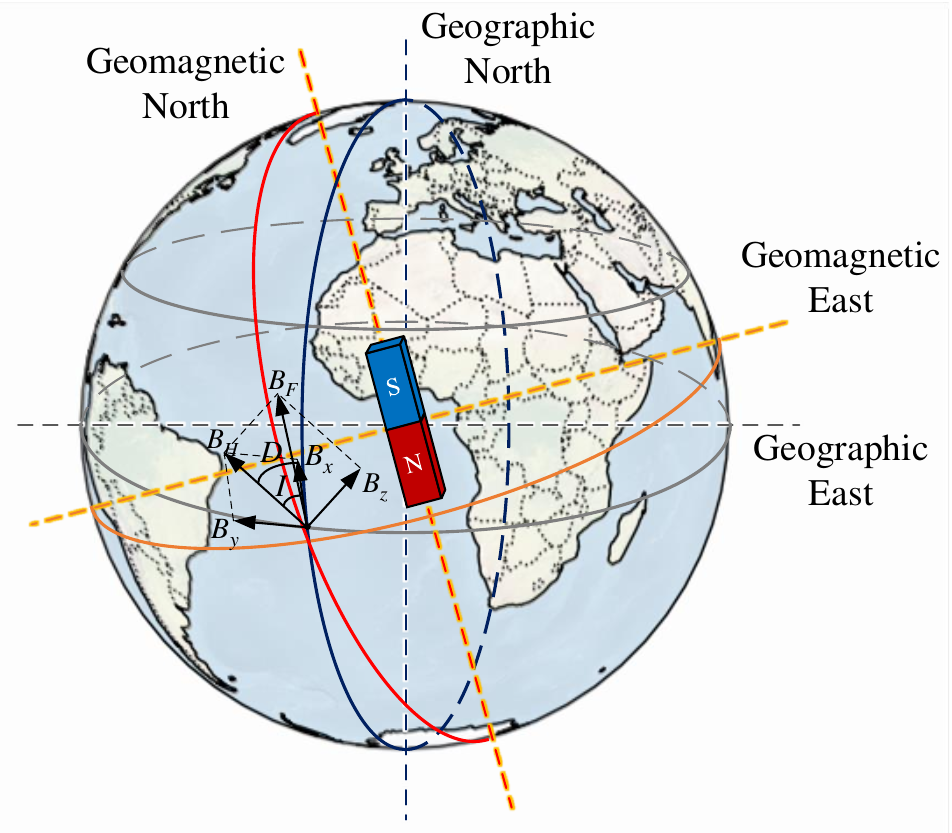}
    \caption{The geomagnetic field of the earth is represented in a spherical coordinate system, the axes of the geomagnetic dipole and the earth's rotation form an angle of approximately 11°. 
    }
    \label{fig:2-1}
\end{figure}

To describe the spatial distribution of the geomagnetic field, the geomagnetic vector is decomposed into geomagnetic parameters \(({B_F}, {B_H}, {B_x}, {B_y}, {B_z}, {D}, {I})\) in the geographic coordinate system as shown in \Cref{fig:2-1}. The total strength of the geomagnetic field is denoted by \({B_F}\), and its projection on the horizontal plane is the horizontal component, denoted by \({B_H}\). \({B_x}\) and \({B_y}\) are the projections of \({B_H}\) in the North direction and East direction, respectively, and \({B_z}\) is the projection of \({B_F}\) in the vertical direction, known as the vertical component. \({D}\) is called the magnetic deviation, which is the angle between \({B_H}\) and \({B_x}\) in the horizontal plane. The angle between \({B_F}\) and the horizontal plane is called magnetic inclination, denoted by \({I}\). The seven elements can be calculated by

\begin{align}
\left\{\begin{array}{l}
B_x=B_F \cos I \cos D \\
B_y=B_F \cos I \sin D \\
B_z=B_F \sin I \\
B_F=\sqrt{B_x^2+B_y^2+B_z^2} \\
B_H=B_F \cos I \\
I=\arccos \left(B_H / B_F\right) \\
D=\arccos \left(B_x / B_H\right)
\end{array}\right..
\label{eq:2-1}
\end{align}

The geomagnetic field is a function of spatial position and time, and various mathematical models have been developed to represent the geomagnetic field. Notable models include the International Geomagnetic Reference Field Model (IGRF), the World Magnetic Model (WMM), and the Enhanced Magnetic Model (EMM). The IGRF is a standard mathematical description of the large-scale structure of the geomagnetic field and its secular variation. It was created by fitting parameters of a mathematical model to measured magnetic field data from surveys, observatories and satellites across the globe \cite{thebault2015international}. The IGRF model spans a considerable time period, making it valuable for interpreting historical data. It is updated every five years to incorporate the most accurate measurements available at the time.

\subsection{Problem Formulation}
\label{Section:2.2}
The goal of geomagnetic navigation is to identify an optimal or suboptimal path from an origin to a destination in a two-dimensional (2D) or three-dimensional (3D) environment through geomagnetic information. Considering that the vehicle relies solely on geomagnetic information, magnetic sensors measure geomagnetic parameters in real-time, and use the disparity between geomagnetic parameters at the present position and the destination to determine the movement distance and yaw angle for movement \cite{watkins1992q}. We regard the vehicle as a particle and only consider the problem in a two-dimensional Cartesian coordinate system, the movement of the vehicle is
\begin{align}
    \left\{ \begin{array}{l}
    {x_{j + 1}} = {x_{j}} + {L_{j+1}}\cos ({\theta _{j+1}})\\
    {y_{j + 1}} = {y_{j}} + {L_{j+1}}\sin ({\theta _{j+1}})
    \end{array} \right.,
\label{eq:2-2}
\end{align}
where \({x_j},{y_j}\) is the coordinate position of the vehicle in time step \(j\), \({L_j}\) and \({\theta _j}\) respectively represent the movement distance and angle of the vehicle in the Cartesian coordinate system. 

In this article, the angular displacement of the vehicle \({\theta _j}\) is defined as the yaw angle \({\psi _j}\) after previous time step. Accordingly, the motion equation of the vehicle is represented as
\begin{align}
    \left\{ \begin{array}{l}
    {x_{j + 1}} = {x_j} + {L_{j+1}}\cos ({\theta _j+\psi _{j+1}})\\
    {y_{j + 1}} = {y_j} + {L_{j+1}}\sin ({\theta _j+\psi _{j+1}})
    \end{array} \right..
\label{eq:2-3}
\end{align}

From a reinforcement learning perspective, a strong and constrained relationship exists between geomagnetic parameters and the navigation path. This enables the agent to develop a "magnetotactic" capability similar to that seen in migrating animals. By leveraging the sensitivity of geomagnetic trends to the destination, the agent establishes a convergence with geomagnetic parameters throughout the navigation process.

The agent taking one or more of the geomagnetic parameters as shown in   \Cref{eq:2-1} at its current position and destination as perceptual information, which can be denoted by an \(i\)-dimensional vector

\begin{align}
\mathbf{B} = \{ {B_1},{B_2}, \ldots ,{B_i}\}.
\label{eq:2-4}
\end{align}

The agent iteratively selects navigation actions including the yaw angle \({\psi _j}\) and the movement distance \({L_j}\) to maximize rewards, guiding the vehicle to converge from the current position to the destination. When the geomagnetic parameters measured at the current position are equal to those at the destination, it signifies that the vehicle has reached the destination and completed the navigation process \cite{liu2013bio}, that is
\begin{align}
    \lim _{k \rightarrow \infty} F(\mathbf{B}, k) \rightarrow 0,
    \label{eq:2-5}
\end{align}
where \(F(\mathbf{B}, j)\) is  the normalized objective function at the \(j\)-th time step, which can be expressed as
\begin{align}
F(\mathbf{B}, j)=\sum_{i=1}^n \frac{f_i^j(\mathbf{B}, j)}{f_i^{\prime}(\mathbf{B}, j)}=\sum_{i=1}^n \frac{\left(\mathbf{B}_i^{T}-\mathbf{B}_i^j\right)^2}{\left(\mathbf{B}_i^{T}-\mathbf{B}_i^0\right)^2},
\label{eq:2-6}
\end{align}
\(\mathbf{B}_i^j\) and  \(\mathbf{B}_i^{T}\) are the current value and the target value of the \(i\)-th geomagnetic parameter, respectively, \(f_i(\mathbf{B}, j)\)  represents the objective sub-function of the \(i\)-th geomagnetic parameter at the \(j\)-th time step. 

In practice, the vehicle is considered to have successfully navigated to the destination when it is within a certain distance. This can be equivalent to the total objective function being less than the set threshold \(\zeta\)
\begin{align}
    F(\mathbf{B}, j)<\zeta.
\label{eq:2-7}
\end{align}

Therefore, we regard the navigation of autonomous mobile vehicles in unknown environments without prior information as a reinforcement learning task. This task involves training the agent to select actions in the navigation space to maximize cumulative rewards and successfully complete the navigation task.

\section{Gradient-Guided TD3 algorithm for Geomagnetic navigation}
\label{Section:3}
Deep reinforcement learning is a branch of machine learning that focuses on how artificial agents make decisions in an environment to maximize cumulative future rewards. In this section, we introduce the Markov decision process (MDP), which serves as the theoretical foundation of DRL, to describe our DRL-based solution. Subsequently, we introduce the selected DRL framework TD3 algorithm, along with the optimized Gradient-Guided TD3 algorithm proposed in this article to address the geomagnetic navigation task.

\subsection{Solution of DRL-Based Geomagnetic Navigation}
\label{Section:3.1}
Deep reinforcement learning is typically described using MDP. Specifically, for the geomagnetic navigation problem, the current position of the vehicle depends only on its previous position and navigation actions. Consequently, we interpret the geomagnetic navigation problem as an MDP, with the definitions of its state, action, and reward outlined below.

The MDP is typically characterized by a 4-tuple \((\mathcal{S}, \mathcal{A},\mathcal{P}, \mathcal{R})\), where \(\mathcal{S}\) represents the state space, comprising a set of environmental states, \(\mathcal{A}\) is the action space, representing a set of available actions, \(\mathcal{P}:\mathcal{S}\times\mathcal{A}\times\mathcal{S}\to [0,1]\) is the dynamics transition function from the current state \(s_j\) to the next state \(s'_j\) under action \(a_j\), and \(\mathcal{R}:\mathcal{S}\times\mathcal{A}\to \mathbb{R}\) is the reward function, providing immediate rewards following state transitions \cite{arulkumaran2017brief}. Details on the MDP for the geomagnetic navigation problem are as follows:

\begin{enumerate}
    \item \textit{State} \(s_j\): The state is composed of information available at each time step \(j\), which consists of real-time geomagnetic parameters measured by magnetic sensors, movement distance, and yaw angle in the previous time step. Considering that any three-dimensional parameters allow us to obtain all geomagnetic parameters. Therefore, in this article, we select only three geomagnetic parameters: the magnetic deviation \({D}\), the magnetic inclination \({I}\), and the horizontal component \({B_H}\). The state \(s_j\) is defined as
    \begin{align}
        s_j=\{D^j,I^j,B_H^j,D^{T},I^{T},B_H^{T},L_j,\psi_j\},
        \label{eq:3-1}
    \end{align}
    where \(\{D^j,I^j,B_H^j\}\) and \(\{D^{T},I^{T},B_H^{T}\}\) respectively denote the three-dimensional geomagnetic parameters measured at the current position and the destination, \(L_j\) and \(\psi_j\) is the movement distance and yaw angle of the vehicle.

    \item \textit{Action} \(a_j\): The agent takes action \(a_j\) after obtaining the state \(s_j\) at time step \(j\). The action \(a_j\) is defined as
    \begin{align}
    a_j=\{L_{j+1},\psi_{j+1}\},
        \label{eq:3-2}
    \end{align}

    It is noteworthy that research in the field of geomagnetic navigation typically considers the yaw angle \(\psi_{j}\) as the output of the navigation algorithm, while the movement distance \(L_{j}\) is usually pre-set to a fixed value. However, the navigation trajectory is influenced by the chosen value of the movement distance. Zhang et al. \cite{zhang2019bio} underscore the complexities associated with determining the movement distance \(L_j\) in long distance navigation. When \(L_j\) is relatively small, substantial changes in gradient suggest a localized influence, making it challenging to represent the overall gradient of the entire area effectively, as \(L_j\) increases, the gradient tends to stabilize. Nevertheless, in geomagnetic navigation, due to the absence of location information, the navigation is deemed successful when the geomagnetic parameters at the current position fulfill the condition in \Cref{eq:2-7}. Setting an excessively large movement distance may result in the solution not converging within the convergence domain.

    \item     \textit{Reward} \(r_j\): We design a composite reward function, which is set as
\begin{align}
    r_j=R_{\text{destination}}^j+R_{\text{proximity}}^j+R_{\text{alignment}}^j,
    \label{eq:3-3}
\end{align}
the reward function consists of three parts, the destination reward \(R_{\text{destination}}^j\) is set to encourage the agent to reach the destination and is represented by a positive value \(\zeta_1\). When the vehicle reaches the destination, the agent acquires the destination reward. The proximity reward \(R_{\text{proximity}}^j\) is defined as
\begin{align}
    R_{\text{proximity}}^j=-\zeta _2(F(\mathbf{B}_j,j) - F(\mathbf{B}_{j-1},j-1)),
    \label{eq:3-5}
\end{align}

To gauge the distance between the current position and the destination using geomagnetic parameters, we employ the objective function \(F(\mathbf{B}_j,j)\) defined in \Cref{eq:2-6}. Rewards are given for actions that move toward the destination, and penalties are imposed for those that move away, based on the comparison of the objective function between the current and previous time steps. We implement a reward decay mechanism for the \(R_{\text{proximity}}^j\), when the current time step exceeds half of the maximum time step \(n_{\text{max}}\), an additional penalty term proportional to the \(R_{\text{proximity}}^j\) and \(n_j\) is applied. \(R_{\text{proximity}}^j\) is set to encourage action that converges geomagnetic parameters from the current position toward the destination between adjacent time steps, prevent the agent from adopting an excessively conservative speed.

The alignment reward \(R_{\text{alignment}}^j\) in this article is shown in \Cref{eq:3-6}:
\begin{align}
        R_{\text{alignment}}^j=\zeta _3(\pi / 4 - |\lambda_j - \lambda'_j|),
    \label{eq:3-6}
\end{align}
where \(\lambda_j\) represents the angular displacement of the vehicle, accumulated from the yaw angle \(\psi_j\) at each time step, and \(\lambda'_j\) represents the theoretical heading angle calculated using the parallel approach based on magnetic gradients, with specific calculations to be introduced in \Cref{Section:3.3}. The \(R_{\text{alignment}}^j\) is designed to constrain the heading deviation. 

\(\zeta_1\), \(\zeta_2\), and \(\zeta_3\) are three adjustable parameters, where \(\zeta_1\) should be set to the maximum. \(\zeta_2\) and \(\zeta_3\) are the weights for two subsidiary rewards. Notably, due to the non-uniform distribution of the magnetic field, a large value of \(\zeta_3\) may decrease the efficiency of geomagnetic navigation in areas with magnetic field anomalies. Therefore, setting the impact of \(R_{\text{alignment}}\) on the training process should not exceed that of \(R_{\text{proximity}}\). We experimented with different values of parameters \(\zeta_1\), \(\zeta_2\), and \(\zeta_3\) when training the TD3 agent. To balance the three goals mentioned above, we set the value of \(\zeta_1\) to 200, \(\zeta_2\) to 10, and \(\zeta_3\) to 3.

    \item The transitions of the geomagnetic multi-parameters convergent space and navigation motive space have been discussed in \Cref{Section:2}. However, the state depends not only on geomagnetic parameters but also on the navigation action at the previous time step. The developed DRL algorithm will learn the correlations from the training data to make optimal decisions.
\end{enumerate}

\subsection{Basic Principle of the TD3 Algorithm}
\label{Section:3.2}
\begin{figure}[htb]
    \centering

			\includegraphics[width=0.75\linewidth]{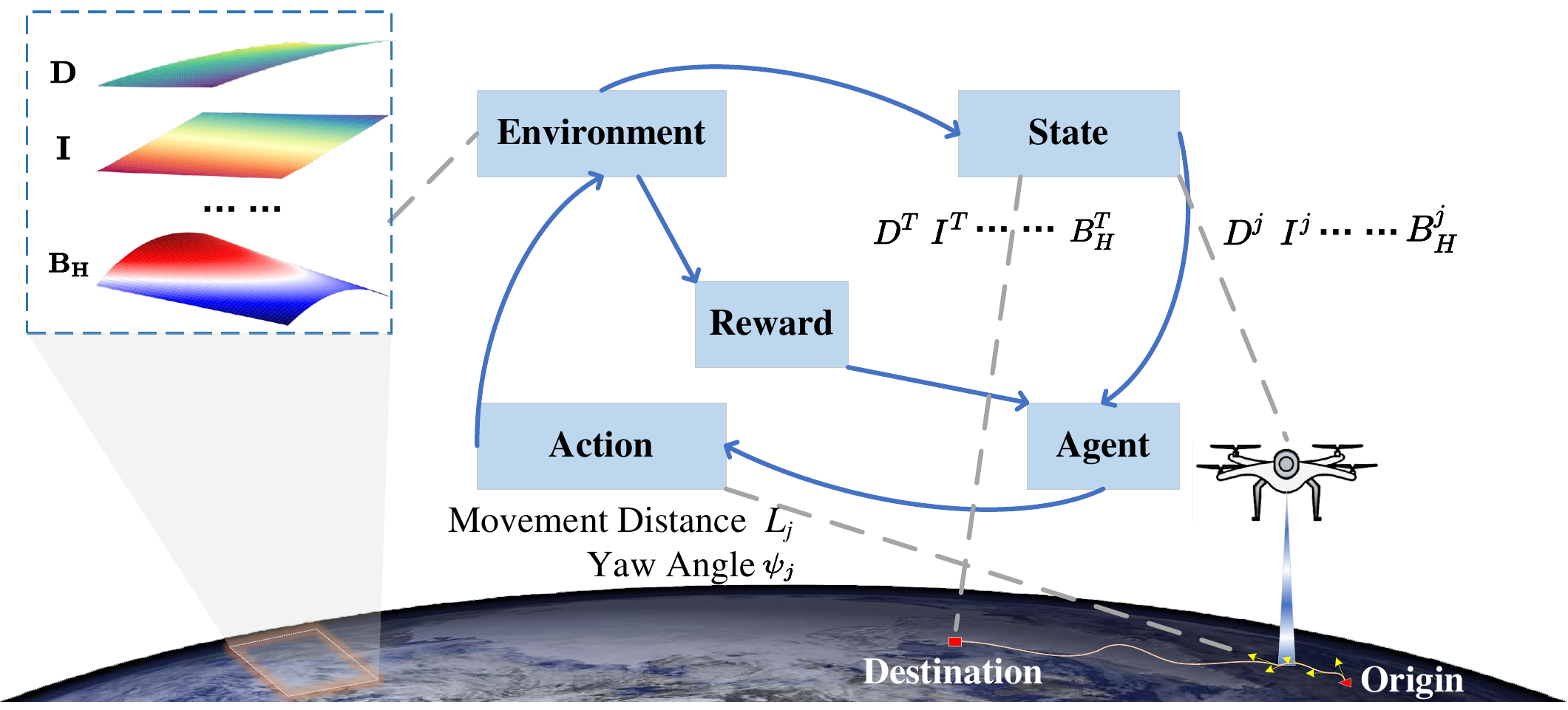}
		
    \caption{Interactions between the DRL agent and the environment in geomgnetic navigation. 
    }
    \label{fig:3-1}
\end{figure}

DRL operates on an iterative 'trial and error' learning process, where the agent learns to maximize rewards through interactions with the environment \cite{tan2019energy}. As illustrated in \Cref{fig:3-1}, the agent selects actions based on the current state, and the environment provides feedback in the form of new state and reward signals, transitioning to the next state. Through this iterative process, the agent converges on an optimal policy that maximizes cumulative rewards within the environment.

In DRL, the network is updated through temporal difference learning, utilizing a secondary frozen target network \(Q_{\theta'}(s,a)\) to preserve an objective \(y\) across multiple updates as
\begin{align}
    y=r+\gamma  Q_{\theta'}({s'},{a'})
    \label{eq:3-7}
\end{align}
\begin{align}
    {a'}\sim\pi_{\phi'}(s'),
    \label{eq:3-8}
\end{align}
where the actions are selected from a target actor network \(\pi_{\phi'}\), \(\gamma\) is the discount factor,  \(s'\) and \(a'\) are the next state and action respectively. The weights of a target network are updated either periodically to precisely match the weights of the current network or by a given proportion \(\tau\) at each time step, as shown by \(\theta '\gets \tau \theta + (1 - \tau) \theta '\), \(\phi '\gets \tau \phi + (1 - \tau) \phi '\). The update can be implemented in an off-policy manner, involving the random sampling of mini-batches of transitions from an experience replay buffer \cite{lin1992self}.

In practice, the issue of value overestimation as a result of function approximation errors is well-studied \cite{silver2014deterministic, van2016deep, fujimoto2018addressing}. The maximum value of the estimated \(Q\) value is typically bigger compared to the real value due to noise in the samples. This noise introduces inaccuracies, thereby reducing the precision of the \(Q\) function. During policy updates, inaccurate estimations can result in error accumulation. These accumulated errors contribute to the overestimation of state values. Consequently, the policy optimization may deviate from the optimal path, impeding algorithm convergence. 

\begin{figure}[htb]
    \centering
    \includegraphics[width=0.98\linewidth]{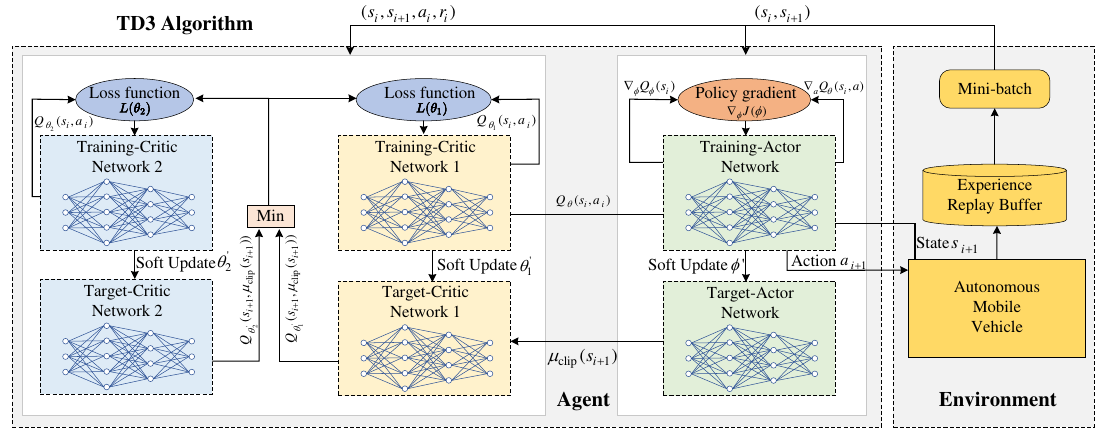}
    \caption{The framework of TD3 algorithms for geomagnetic navigation.
    }
    \label{fig:3-2}
\end{figure}

For the overestimation problem of \(Q\) value,  the TD3 algorithm is utilized to realize the navigation strategy of vehicles.  TD3 belongs to actor-critic algorithms, it mitigates the overestimation problem of \(Q\) value by integrating the DDPG and DDQN. The architecture of TD3 is shown in \Cref{fig:3-2},  involving the copies of neural networks for one actor network and two critic networks as target networks. Two critic networks are employed to evaluate the \(Q\) value, and the smaller one is selected to update the target \(Q\) value as shown in \Cref{eq:3-9}. The loss function is defined as the squared difference between the selected target \(Q\) value and the output of the neural network, as expressed in \Cref{eq:3-10}:
\begin{align}
    y=r+\gamma \min _{i=1,2} Q_{\theta_i^{\prime}}(s', a')
    \label{eq:3-9}
\end{align}
\begin{align}
    L(\theta_i)=E[(\theta_i(s, a)-y)^2].
    \label{eq:3-10}
\end{align}

In addition, TD3 introduces the regularization of parameter updates to reduce the deviation generated by the estimation of the \(Q\) function. The noise \(\varepsilon\) is added to the target action as a regularization in \Cref{eq:3-11} to facilitate a smoother update process and mitigate the risk of overfitting.
\begin{align}
    a'(s')=\text{clip}(\pi_{\phi'}(s')+\varepsilon,a_{\text{Low}},a_{\text{High}}),
    \label{eq:3-11}
\end{align}

\begin{algorithm}
\caption{Twin Delayed DDPG for the Geomagnetic Navigation}
\label{Algorithm:4}
\SetAlgoLined
\KwIn{initial parameters of critic networks \(\theta_1, \theta_2\), parameters of actor networks \(\phi\),  discount factor \(\gamma \), size of sample batches \(S\), empty replay buffer \(\mathcal{D}\) and max capacity \(N\)}

Set target parameters equal to main parameters \(\theta'_1\gets \theta_1\), \(\theta'_2\gets \theta_2\), \(\phi' \gets \phi\)

\Repeat {\textup{convergence}}{
    Observe state \(s\) and select action by \Cref{eq:3-11};
    Execute \(a\) in the environment\;
    Observe next state \(s'\), reward \(r\), and done signal \(d\) to indicate whether \(s'\) is terminal\;
    Store \((s,a,r,s',d)\) in replay buffer \(\mathcal{D}\)\;
    If \(s'\) is terminal, reset environment state\;
    \If {\textup{it's time to update}}{
        \For{\(j\) \textup{in range (however many update)}}{
            Randomly sample a batch of transitions, \(B=\{(s,a,r,s',d)\}\) from \(\mathcal{D}\)\;
            Compute target actions by \Cref{eq:3-11}\;
            Compute targets \(y\) by \Cref{eq:3-9}\;
            Update \(\theta_i:\\ \theta_i \gets \arg \min _{\theta_i} |B|^{-1} \sum\left(y-Q_{\theta_i}(s, a)\right)^2\) \;
            Update \(\phi\) by the deterministic policy gradient:\(\\ \nabla _\phi J(\phi)=B^{-1}\sum \nabla_aQ_{\theta}(s,a)|_{a=\pi_\phi(s)}\nabla _\phi\pi _\phi(s)\)\;
            \If {j \(\textup{mod}\ \texttt{policy\_delay}=0\)}{
                Update target networks using \(\\\theta'_i \gets \tau \theta_i + (1-\tau)\theta'_i\\
    \phi' \gets \tau \phi + (1-\tau)\phi'\)
            }
        }
    }
}
\end{algorithm}

\noindent where \(\varepsilon \in \text{clip}(\mathbb{N}(0,\sigma ),-c,c)\), \(\varepsilon\) represents the noise, \(\sigma\) denotes the standard deviation of the noise,  \(\mathbb{N}\) refers to the standard normal distribution, and \(c\) defines the noise bound. The detailed pseudocode is provided in \Cref{Algorithm:4}.

\subsection{Optimized Gradient-Guided TD3 Method}
\label{Section:3.3}
For long-distance geomagnetic navigation, the smoothness of the navigation trajectory is crucial as it reflects the efficiency of the navigation strategy. To efficiently guide the agent in navigating through an unfamiliar environment, we draw inspiration from the bionic navigation approach proposed by Wang et al. \cite{wang2018geomagnetic}. The gradient of geomagnetic parameters is introduced to calculate the theoretical heading angle approximately pointing toward the destination by parallel approach. Considering the overall stability of geomagnetic field distribution, encouraging the agent to choose navigation actions that align closely with the theoretical heading angle can enhance exploration efficiency in practice.

\begin{figure}[h]
    \centering
    \includegraphics[width=0.65\linewidth]{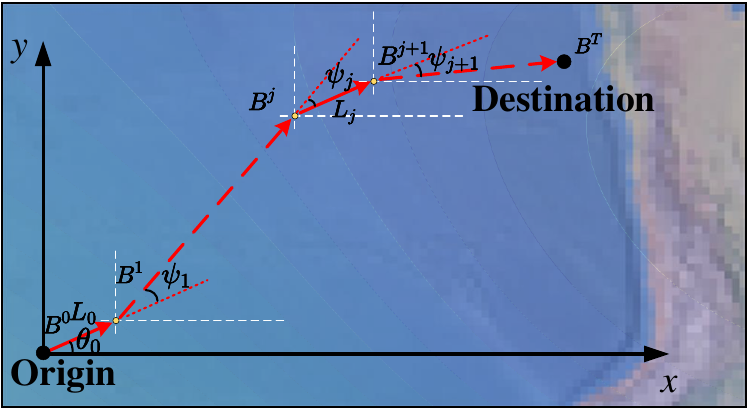}
    \caption{Deduction of the bio-inspired geomagnetic navigation.
    }
    \label{fig:3-3}
\end{figure}
As depicted in \Cref{fig:3-3}, \(B^0\) denotes the vector of geomagnetic parameters at the origin,  \(B^T\) denotes the vector at the destination, and the vector at the \(j\)-th step is denoted as \(B^T\). The parallel approach allows different geomagnetic parameters to approach their respective target \(B_i^T\) with the same ratio as follows:
\begin{align}
(B_i^{j+1}-B_i^j)\propto (B_i^T-B_i^j).
\label{eq:3-12}
\end{align}

Projecting vectors \((B_i^{j+1}-B_i^j)\) and \((B_i^T-B_i^j)\) onto the geographical coordinate system, the following results can be obtained:
\begin{align}
\frac{B_{i_1}^{j+1}-B_{i_1}^j}{B_{i_1}^T-B_{i_1}^j} =\frac{B_{i_2}^{j+1}-B_{i_2}^j}{B_{i_2}^T-B_{i_2}^j},
\label{eq:3-13}
\end{align}
where geomagnetic parameters at two adjacent time steps satisfy the relationship as
\begin{align}
\left\{\begin{matrix}B_{i_1}^{j+1}=B_{i_1}^j + g_{{i_1},x}^j\cdot \cos \theta_k + g_{{i_1},y}^j\cdot \sin \theta_k 
 \\B_{i_2}^{j+1}=B_{i_2}^j + g_{{i_2},x}^j\cdot \cos \theta_k + g_{{i_2},y}^j\cdot \sin \theta_k 
\end{matrix}\right.,
\label{eq:b+14}
\end{align}
where \(g_{{i_1},x}^j\), \(g_{{i_1},y}^j\), \(g_{{i_2},x}^j\), \(g_{{i_2},y}^j\) are the gradients of \(B_{i_1}^j\) and \(B_{i_2}^j\). By substituting \Cref{eq:b+14} into \Cref{eq:3-13}, the theoretical heading angle can be calculated as
\begin{align}
\lambda'_j = \arctan(\frac{(B_{i_1}^j-B_{i_1}^T) \cdot g_{i_2,x} - (B_{i_2}^j-B_{i_2}^T) \cdot g_{i_1,x}}{(B_{i_2}^j-B_{i_2}^T) \cdot g_{i_1,y} - (B_{i_1}^j-B_{i_1}^T) \cdot g_{i_2,y}}).
\label{eq:3-15}
\end{align}

The theoretical heading angle \(\lambda'_j\) can be estimated from previous calculations. However, while the geomagnetic field distribution is generally stable, magnetic anomalies can cause inaccuracies in gradient calculations. This makes it problematic to directly use the theoretical heading angle in navigation tasks. As a more effective strategy, we set a low-weight alignment reward to encourage the yaw angle to closely match the theoretical heading angle.

\section{Experiments}
\label{Section:4}

In this section, we outline the experimental setup, detailing the simulation environment, evaluation metrics, and implementation procedures. We then conduct an in-depth analysis of the algorithm's performance in geomagnetic navigation, using visual comparisons of long-distance navigation trajectories to emphasize effectiveness of algorithms. Finally, we present quantitative results to offer a thorough understanding of the performance metrics.

\subsection{Experimental Setup}
\label{Section:4.1}

\subsubsection{Simulation Environment} 
\begin{figure}[htb]
    \centering
    \includegraphics[width=0.75\linewidth]{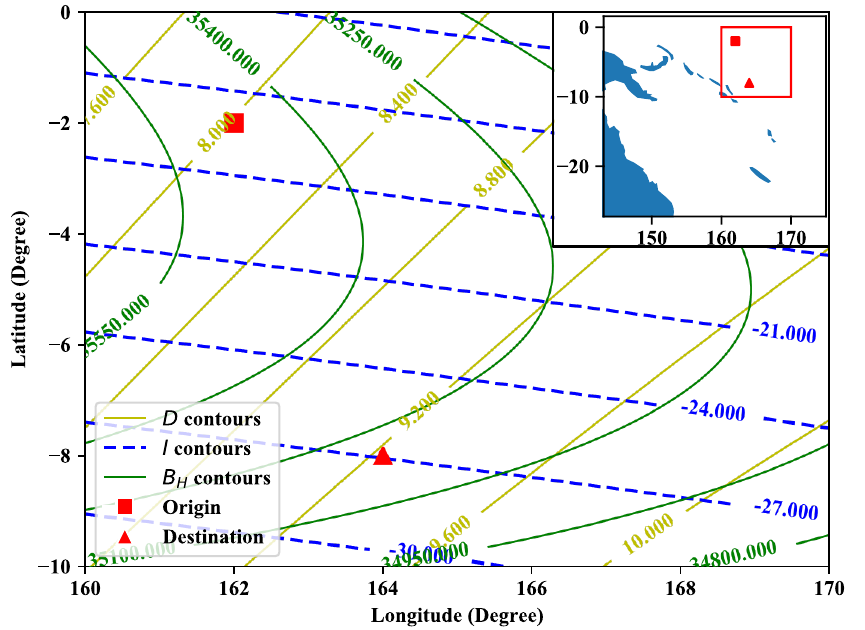}
    \caption{Contour maps demonstrating magnetic deviation \({D}\), magnetic inclination \({I}\), and the horizontal component \({B_H}\) as derived from the IGRF model, within the region spanning from \((10^\circ S,160^\circ E)\) to \((0^\circ N,170^\circ E)\).}
    \label{fig:4-1}
\end{figure}

Geomagnetic parameters were sourced from the IGRF model. The region selected spans from \((10^\circ S,160^\circ E)\) to \((0^\circ N,170^\circ E\)), encompassing an area of ocean north of Australia, as shown in \Cref{fig:4-1}, the yellow, blue dashed, and green lines represent the contours of \(D\),  \(I\), and \(B_H\), respectively. The red square marks the starting position, while the red triangle indicates the destination.

In \Cref{fig:4-1}, we randomly simulated a navigation task with the origin at \((2^\circ S, 162^\circ E)\) near Nauru, where the geomagnetic parameters are \(D=8.019^\circ\), \(I=-16.150^\circ\), and \(B_H=35467.990\text{nT}\). The destination is set at \((8^\circ S, 164^\circ E)\) near the Solomon Islands, with geomagnetic parameters of \(D=9.228^\circ\), \(I=-26.923^\circ\), and \(B_H=35199.415\text{nT}\). The vehicle starts at the origin, using the geomagnetic parameters of the current position and destination as perceptual information. It navigates based on the selected heading angle and movement distance determined by the navigation algorithms.

\subsubsection{Baselines} 
To comprehensively evaluate the overall performance of the proposed Gradient-Guided TD3 algorithm in unknown environments, we selected four popular methods as baselines: TD3 \cite{fujimoto2018addressing}, genetic algorithm (GA) \cite{holland1975adaptation}, particle swarm algorithm (PSO) \cite{kennedy1995particle}, and artificial fish swarm algorithm (AFSA) \cite{yazdani2011fuzzy}. These baselines highlight two distinct strategies for geomagnetic navigation without prior magnetic maps: TD3 serves as the foundation for the proposed Gradient-Guided TD3 and represents reinforcement learning-based methods, while GA, PSO, and AFSA exemplify metaheuristic approaches commonly employed in bionic geomagnetic navigation.
    
\subsubsection{Evaluation Metrics} 
Our experiments adopted a set of evaluation metrics to assess algorithm performance. Initially, we report four key metrics—success rate (SR)  \cite{sang2022novel}, trajectory length (TL) \cite{lin2023towards}, success weighted by path length (SPL)  \cite{batra2020objectnav}, and total navigation time (TNT) \cite{zhang2022ipaprec} —to provide an overall evaluation of each algorithm’s success rate and navigation efficiency. SR measures the ratio of successful episodes, TL calculates the average length of the trajectories, SPL assesses the efficiency relative to the shortest path, and TNT indicates the average time taken to complete the navigation, with lower values reflecting greater efficiency.

Following this, we further analyzed the algorithms using four additional metrics—path smoothness \cite{hidalgo2017solving}, mean absolute error (MAE) of heading deviation \cite{sabet2017low}, root mean square error (RMSE) of heading deviation \cite{sabet2017low}, and navigation error (NE) \cite{lin2023towards}. path smoothness evaluates the continuity of the agent’s path, MAE and RMSE capture the accuracy and variance in heading angles, and NE measures the final distance between the agent and the target. These metrics were statistically analyzed and presented using box plots to offer deeper insights into task-specific performance.

\begin{table}[ht]
  \caption{Hyperparameter Configuration in the Simulations}
  \centering
  \begin{tabular}{cclcl}
    \toprule%
    \textbf{Hyper Parameter} & \textbf{Value} \\
    \midrule
    Learning rate (\(\lambda _{\text{critic}}, \lambda _{\text{actor}}\)) & 0.001, 0.001 \\
    Policy update delay factor (\(N_{\text{pud}}\)) & 2 \\
    Discounted factor (\(\gamma\)) & 0.995 \\
    Soft target update factor (\(\tau\)) & 0.005 \\
    Capacities of replay buffers (\(M\)) & 50000 \\
    Max episode (\(T\)) & 50 \\
    Size of mini-batch  (\(m\)) & 256 \\
    Number of filter  (\(f_1^C, f_2^C, f_3^C\)) & 512, 512, 512 \\
    Number of episodes  (\(\text{epi}\)) & 20000 \\
    Scale of noise  (\(\alpha _{\text{exploration}}, \alpha _{\text{policy}}\)) & 0.2, 0.1 \\
    \bottomrule
  \end{tabular}
  \label{tab:4-1}
\end{table}

\subsubsection{Implementation Details} 
To select the movement distance \(L_j\) and yaw angle \(\psi_j\) for navigation based on the predicted actions \(a'_j\) at each time step, constraints are applied: the movement distance is limited to a range of 0 to 50 km, and the yaw angle is restricted to \(-\pi/2\) to \(\pi/2\). Both the action vector \(a_j\) and state vector \(s_j\) undergo min-max normalization. The hyperparameters for the proposed Gradient-Guided TD3 algorithm are outlined in \Cref{tab:4-1}. For a fair comparison, 100 navigation tasks were randomly generated within the selected region, with straight-line distances ranging from 300 to 500 km. Each baseline model then sequentially executed all assigned navigation tasks.

\subsection{Analysis of Training Efficiency}
\label{Section:4.2}

\begin{figure}[htb]
    \centering
    \includegraphics[width=0.65\linewidth]{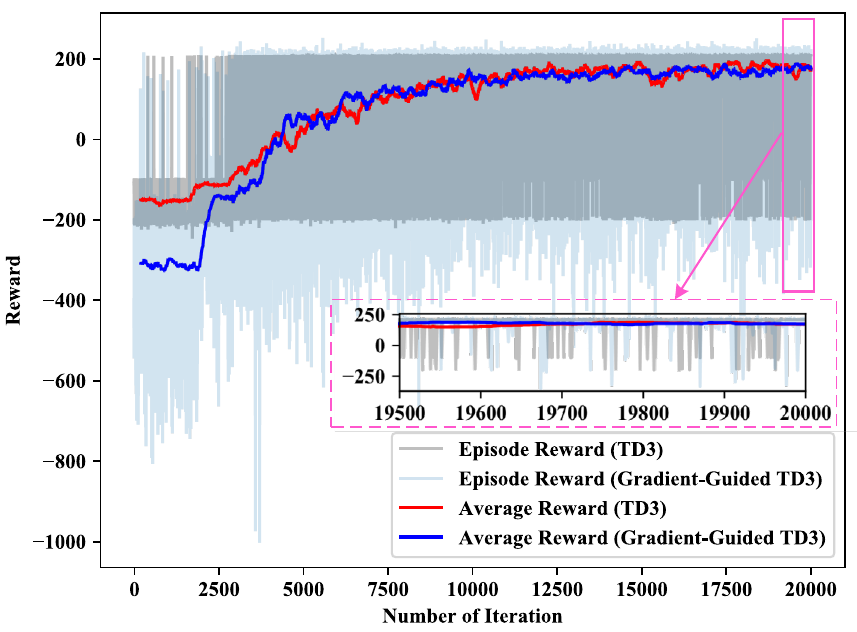}
    \caption{Reward curves of the DRL agent over 20,000 training iterations.}
    \label{fig:4-2}
\end{figure}

\Cref{fig:4-2} presents the cumulative reward curves for the two DRL-based models following 20,000 training iterations in the simulation environment. The gray and light blue lines show the cumulative rewards for each episode. Given the typical fluctuations in episode rewards during reinforcement learning, especially in complex tasks with dense reward structures, it can be challenging to accurately gauge the performance of agent. To address this, we utilize a sliding window technique to smooth the episode rewards. The red and blue lines depict the average rewards, \(\bar{r} _i\), calculated as \(\bar{r} _i=\frac{1}{N} {\textstyle \sum_{i=j-N+1}^{j+1}r_j} \), where \(N\) is the sliding window size (set to 200) and \(r_j\) represents the reward for the \(j\)-th episode during training.

As the DRL-based model continues to train with replay buffer samples, it progressively improves its ability to navigate toward the destination using geomagnetic information, even with random origin and destination points. This improvement is reflected in the steady increase in average rewards as the number of iterations grows. The proposed Gradient-Guided TD3 model exhibits more pronounced fluctuations in the reward curve during the initial training phase due to the added alignment reward \(R_{\text{alignment}}\). However, with sufficient training, the Gradient-Guided TD3 model gradually stabilizes.

\begin{figure}[htb]
    \centering
    \includegraphics[width=0.65\linewidth]{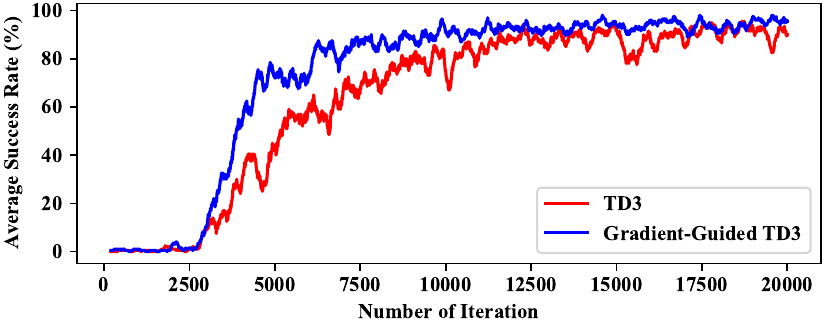}
    \caption{Average success rate of the trained DRL models over 20,000 training iterations.}
    \label{fig:4-3}
\end{figure}

Moreover, \Cref{fig:4-3} presents the average success rate of the DRL models across 20,000 training iterations. It shows the success rate based on the 200 most recent training tasks for each iteration. Both models experience an increase in success rate around the 2,900-\textit{th} iteration. However, the Gradient-Guided TD3 developed in this article demonstrates a more rapid increase in success rate and exhibits less fluctuation. This suggests that the Gradient-Guided TD3 is more effective at learning robust navigation strategies, achieving stability more quickly and consistently compared to the TD3 model.

\subsection{Analysis of Navigation Trajectory}
\label{Section:4.3}

\begin{figure}[htb]
    \centering
    \includegraphics[width=0.85\linewidth]{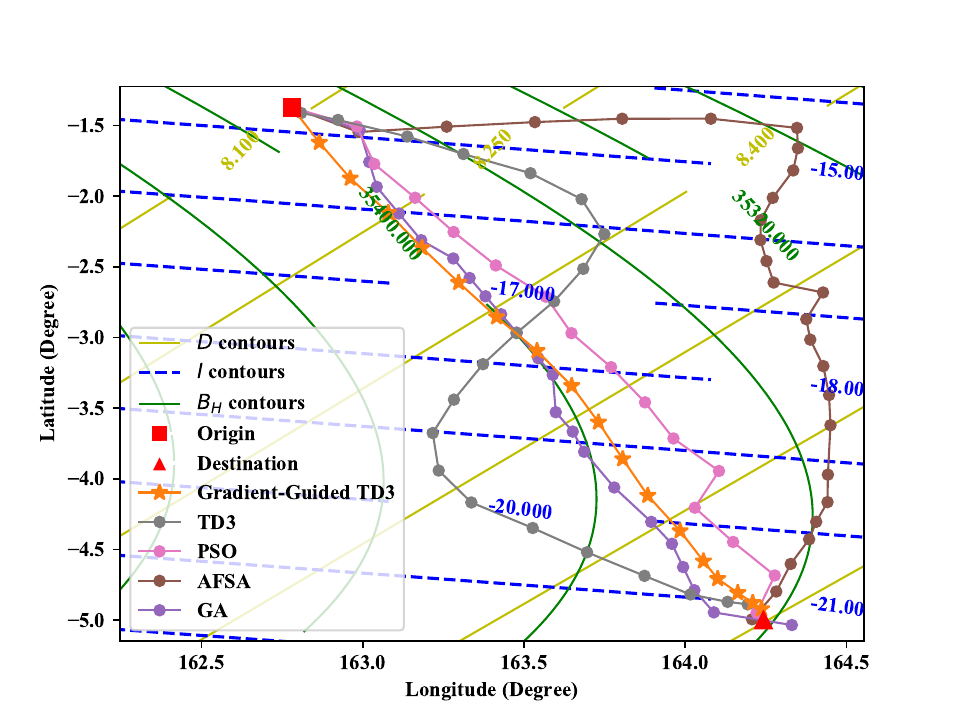}
    \caption{Comparison of the navigation trajectories under different algorithms.}
    \label{fig:4-4}
\end{figure}

The navigation trajectories generated by different algorithms for a typical navigation task are shown in \Cref{fig:4-4}, the trajectories produced by both DRL algorithms are noticeably smoother overall compared to those generated by the metaheuristic algorithms. This disparity arises because metaheuristic algorithms rely heavily on random search strategies, leading to less predictable and more erratic trajectories. On the other hand, DRL algorithms benefit from a training phase that builds a strong association between geomagnetic states and navigation actions, resulting in more stable and coherent trajectories.

Moreover, the enhanced Gradient-Guided TD3 algorithm shows superior linearity in its trajectories compared to the standard TD3. This improvement highlights the advantage of using real-time geomagnetic gradient information to calculate theoretical heading angles, which significantly boosts the navigation algorithm’s efficiency. By incorporating dynamic geomagnetic features, the Gradient-Guided TD3 algorithm adapts more effectively to environmental changes, producing trajectories with improved straight-line characteristics.

\begin{figure}[htbp]
\centering
\subfigure[]
{
    \begin{minipage}{1\linewidth}
        \centering
        \includegraphics[scale=0.85]{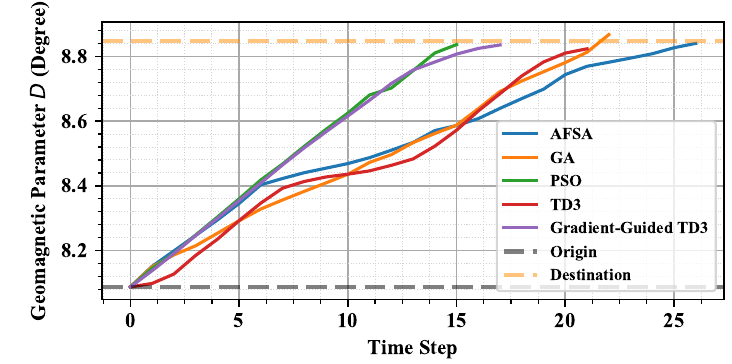}
    \end{minipage}
    \label{fig:4-5-1}
}

\subfigure[]
{
    \begin{minipage}{1\linewidth}
        \centering
        \includegraphics[scale=0.85]{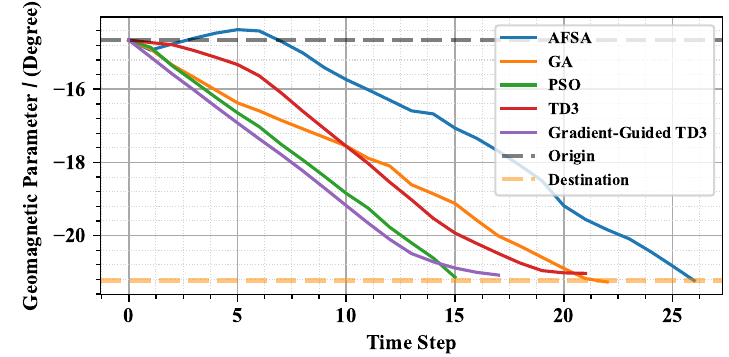}
    \end{minipage}
    \label{fig:4-5-2}
}

\subfigure[]
{
    \begin{minipage}{1\linewidth}
        \centering
        \includegraphics[scale=0.85]{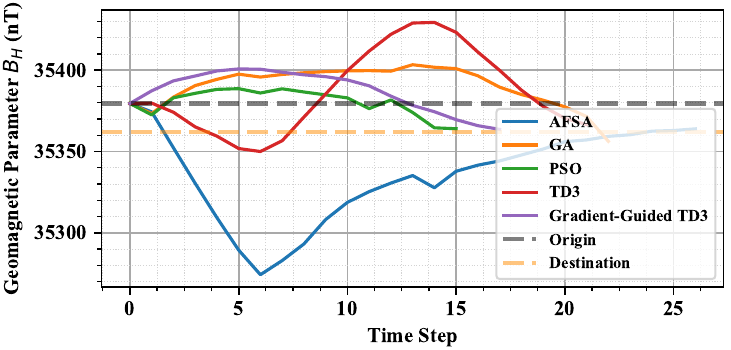}
    \end{minipage}
    \label{fig:4-5-3}
}

\subfigure[]
{
    \begin{minipage}{1\linewidth}
        \centering
        \includegraphics[scale=0.85]{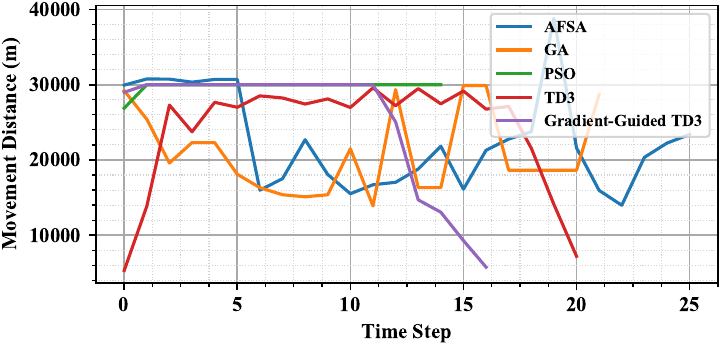}
    \end{minipage}
    \label{fig:4-5-4}
}

\caption{Convergence curve of (a) magnetic deviation \(D\), (b) magnetic inclination \(I\), (c) geomagnetic horizontal component \(B_H\), and (d) movement distance \(L\) across different algorithms.}
\label{fig:4-5}
\end{figure}

The convergence curves of the three geomagnetic parameters are compared between the proposed Gradient-Guided TD3 and baseline methods in \Cref{fig:4-5}, These curves demonstrate how the geomagnetic parameters gradually converge from their initial values towards the target values over time. The results indicate that PSO and the proposed Gradient-Guided TD3 exhibit the fastest convergence, reflecting the highest navigation efficiency. TD3 and GA follow in terms of convergence, while AFSA shows the slowest convergence. The variations in geomagnetic parameters \(D\) and \(I\) between the origin and destination are relatively uniform, whereas \(B_H\) presents non-uniform changes. This non-uniformity introduces deviations in the theoretical heading angles calculated based on the geomagnetic gradient. Further analysis in \Cref{fig:4-5-3} reveals that during navigation, the vehicle, influenced by the non-uniform geomagnetic field strength, does not mechanically follow the gradient descent direction calculated using the parallel approach. Instead, it intelligently identifies a trajectory that is closer to the destination. This highlights the ability of Gradient-Guided TD3 to effectively integrate TD3 with the parallel approach, showcasing its adaptability in navigating through non-uniform geomagnetic field distributions.

\Cref{fig:4-5-4} shows the relationship between movement distance and time step across different methods. Both DRL models generally opt for larger movement distances when far from the destination, shifting to smaller distances as they approach the destination. This behavior is characteristic of geomagnetic navigation, which depends on meeting the threshold condition in \Cref{eq:2-7}. This threshold indicates sufficient proximity to the target without needing additional localization data. Choosing smaller movement distances near the destination helps guide the vehicle into the solution space defined by \Cref{eq:2-7}, reducing the risk of overshooting. This adaptability demonstrates the strength of DRL models in adjusting navigation strategies based on proximity, addressing uncertainties in geomagnetic navigation, and proving their effectiveness in real-world scenarios.

\subsection{Qualitative Results of Navigation in Unknown Environment}
\label{Section:4.4}

To evaluate the accuracy and generalizability of the proposed algorithm in navigating unknown environments, 100 random sets of origins and destinations were independently generated within the selected simulation region. The performance comparison of the different methods across various evaluation metrics is summarized in \Cref{tab:4-2}.

\begin{table}[ht]
  \caption{Performance Comparison of Different Algorithms over the Selected Evaluation Metrics.}
  \centering
  \begin{tabular}{cclcl}
    \toprule
    \textbf{Metrics}& \textbf{SR} (\%) & \textbf{TL} (m) & \textbf{SPL} (\%) & \textbf{TNT} (step)\\
    \midrule
    Gradient-Guided TD3 & \textbf{97} & 445564.313 & \textbf{88.4119} & 21.757 \\
    TD3 & 89 & 511983.904 & 73.6277 & 22.093 \\
    GA & 51 & 443360.591 & 47.0986 & 22.534 \\
    PSO & 29 & \textbf{442495.161} & 27.0941 & \textbf{21.031} \\
    AFSA & 77 & 515649.380 & 63.6267 & 35.560 \\
    \bottomrule
  \end{tabular}
  \label{tab:4-2}
\end{table}

The proposed Gradient-Guided TD3 achieves SR of 97\% and SPL of 88.41\%, outperforming all other methods. This strong success rate highlights the robustness of algorithms in navigating unknown environments. In contrast, PSO and GA show much lower success rates (29\% and 51\%, respectively), indicating difficulties in adapting to unfamiliar settings.

TNT and TL reveal efficiency insights. Although PSO has the lowest TNT, Gradient-Guided TD3 maintains competitive efficiency, balancing efficiency and success rate.

\begin{figure}[htbp]
\centering
\subfigure[]
{
    \begin{minipage}{0.45\linewidth}
        \centering
        \includegraphics[scale=0.6]{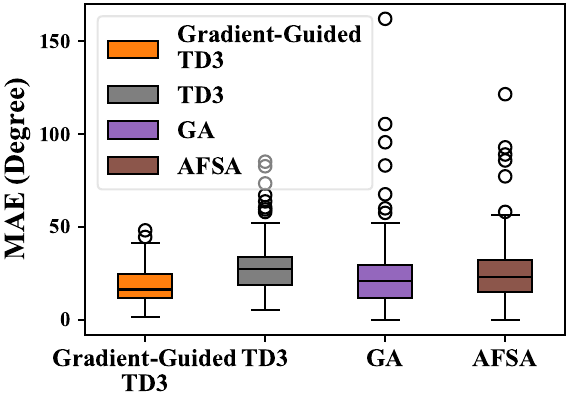}
    \end{minipage}
    \label{fig:4-6-1}
}
\subfigure[]
{
    \begin{minipage}{0.45\linewidth}
        \centering
        \includegraphics[scale=0.6]{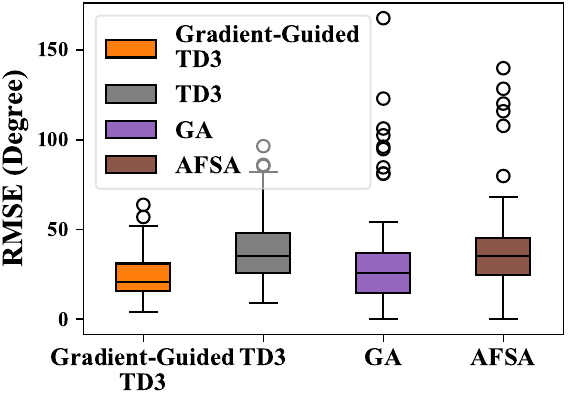}
    \end{minipage}
    \label{fig:4-6-2}
}
\subfigure[]
{
    \begin{minipage}{0.45\linewidth}
        \centering
        \includegraphics[scale=0.6]{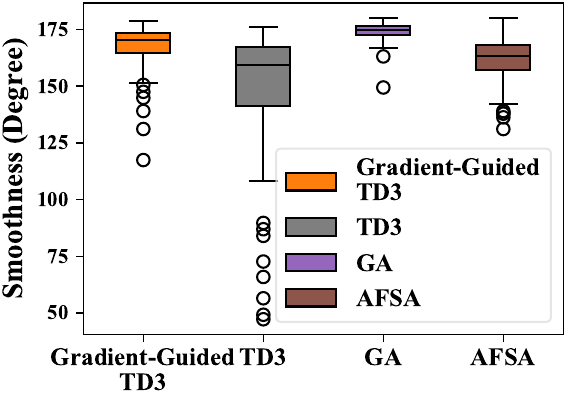}
    \end{minipage}
    \label{fig:4-6-3}
}
\subfigure[]
{
    \begin{minipage}{0.45\linewidth}
        \centering
        \includegraphics[scale=0.6]{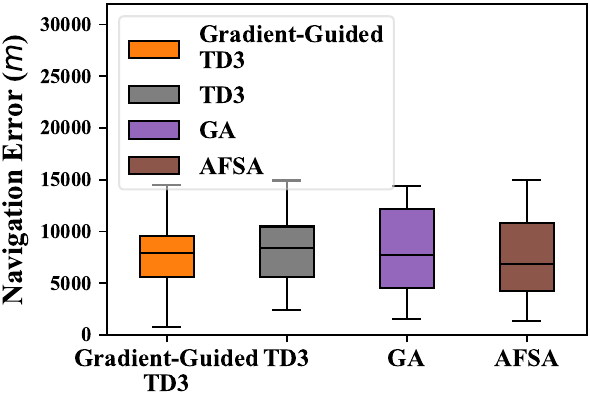}
    \end{minipage}
    \label{fig:4-6-4}
}

\caption{Box Plot of (a) MAE of heading deviation, (b) RMSE of heading deviation,  (c) path smoothness, and (d) NE under algorithms.
}
\label{fig:4-6}
\end{figure}

For a deeper analysis, we examined successful navigation trajectories using box plots in \Cref{fig:4-6}. PSO was excluded due to its low success rate, allowing us to focus on the completed trajectories of other methods.

Examining the RMSE and MAE of heading deviation in \Cref{fig:4-6-2} and \Cref{fig:4-6-1} underscores the exceptional performance of the Gradient-Guided TD3 method. It achieves the smallest mean errors and fewer outliers, demonstrating its ability to maintain stable navigation with minimal deviation, a testament to the adaptive learning of reinforcement learning.

Regarding the smoothness indicator in \Cref{fig:4-6-3}, both the proposed method and GA show commendable results, with average angles over 170 degrees, indicating smoother trajectories. In contrast, TD3 and AFSA struggle more with complex environments, consistent with the trajectory patterns in \Cref{fig:4-4}, where DRL methods show greater adaptability.

Focusing on NE in \Cref{fig:4-6-4}, which measures the distance from the stopping point to the destination, we see that DRL-based methods, including TD3 and Gradient-Guided TD3, reduce movement distance near the target, contributing to more accurate navigation. The box plots confirm that DRL methods offer superior accuracy in long-distance navigation compared to metaheuristic algorithms.

In synthesizing these findings, it becomes evident that the proposed Gradient-Guided TD3 method excels in navigating through unknown environments, leveraging its reinforcement learning foundations to achieve a harmonious balance between precision, stability, and adaptability. The results not only substantiate the advantages of DRL-based approaches but also shed light on the nuanced challenges faced by metaheuristic algorithms in the context of dynamic and unfamiliar environments.

\section{Conclusion}
\label{Section:5}
We introduced Gradient-Guided TD3, a DRL-based algorithm for geomagnetic navigation in unknown and GNSS-denied environments, facilitating long-distance navigation based on geomagnetic information. The developed algorithm efficiently explores geomagnetic features during the navigation progress and guides the carrier toward the destination. By integrating a parallel approach with the exploration of the DRL agent, the developed approach enhances the navigation efficiency and avoids excessive reliance on the theoretical heading angle calculation, ensuring robust long-distance navigation in non-uniform magnetic field distributions.

In essence, the significance of long-distance navigation in unknown environments lies in its potential to broaden the scope of autonomous systems, making them more versatile and applicable across diverse real-world situations. The ability to navigate autonomously over long distances promotes the autonomy, flexibility, and overall utility of navigation systems in scenarios where prior knowledge is limited or unavailable. The insights gained by this study underscore the significance of adaptive, learning-based approaches in the field of autonomous navigation, pave the way for further refinements in algorithmic approaches, and open avenues for exploring the intersection of artificial intelligence and navigation in challenging environments.

\bibliographystyle{IEEEtran}
\bibliography{Reference}

\end{document}